\documentclass[12pt]{article}
\usepackage[utf8]{inputenc}
\usepackage[svgnames]{xcolor}
\usepackage{jmlr2e}
\usepackage[a4paper, margin=1.2cm, bottom=1in]{geometry}
\usepackage{tikz}
\usetikzlibrary{calc,arrows}
\usepackage[framemethod=tikz]{mdframed}
\usepackage{amsmath}
\usepackage{bm}
\usepackage{amssymb}
\usepackage{tikz}
\usepackage{multirow}
\usepackage{graphicx}
\usepackage{booktabs}
\usepackage{array}
\usepackage{pifont}
\usepackage{natbib}
\usepackage{stackengine}

\colorlet{questionColor}{DarkOrange}
\colorlet{exampleColor}{Silver}
\colorlet{answerColor}{LimeGreen}
\colorlet{commentColor}{DodgerBlue}
\colorlet{remarkColor}{FireBrick}
\colorlet{proofColor}{MediumPurple}

\def\questionSymbol{{\textbf{Q}}}
\def\exampleSymbol{{\textbf{E}}}
\def\answerSymbol{{\textbf{A}}}
\def\commentSymbol{{\textbf{C}}}
\def\remarkSymbol{{\textbf{R}}}
\def\proofSymbol{{\textbf{P}}}

\newcommand{\mcrot}[4]{\multicolumn{#1}{#2}{\rlap{\rotatebox{#3}{#4}~}}} 

\let\svendmdframed\endmdframed
\makeatletter
\def\endmdframed{\svendmdframed\unskip}
\makeatother

\def\delequal{\mathrel{\ensurestackMath{\stackon[1pt]{=}{\scriptstyle\Delta}}}}

\tikzset{commentsymbol/.style={rectangle, draw=commentColor, text=commentColor, fill=white, scale=1, overlay}}
\tikzset{examplesymbol/.style={rectangle, draw=exampleColor, text=exampleColor, fill=white, scale=1, overlay}}
\tikzset{answersymbol/.style={rectangle, draw=answerColor, text=answerColor, fill=white, scale=1, overlay}}
\tikzset{questionsymbol/.style={rectangle, draw=questionColor, text=questionColor, fill=white, scale=1, overlay}}
\tikzset{remarksymbol/.style={rectangle, draw=remarkColor, text=remarkColor, fill=white, scale=1, overlay}}
\tikzset{proofsymbol/.style={rectangle, draw=proofColor, text=proofColor, fill=white, scale=1, overlay}}

\mdfdefinestyle{answer}{
hidealllines=true, leftline=true, skipabove=0pt, skipbelow=0pt, innertopmargin=0.7em, innerbottommargin=0.7em,
innerrightmargin=0.7em, rightmargin=0.7em, innerleftmargin=1.4em, leftmargin=0.7em, middlelinewidth=.2em,
linecolor=answerColor, fontcolor=black, backgroundcolor=answerColor!15,
firstextra={\path let \p1=(P), \p2=(O) in (\x2,0.5 * \y1) node[answersymbol] {\answerSymbol};},
secondextra={\path let \p1=(P), \p2=(O) in (\x2,0.5 * \y1) node[answersymbol] {\answerSymbol};},
middleextra={\path let \p1=(P), \p2=(O) in (\x2,0.5 * \y1) node[answersymbol] {\answerSymbol};},
singleextra={\path let \p1=(P), \p2=(O) in (\x2,0.5 * \y1) node[answersymbol] {\answerSymbol};}
}

\mdfdefinestyle{comment}{
hidealllines=true, leftline=true, skipabove=0pt, skipbelow=0pt, innertopmargin=0.7em, innerbottommargin=0.7em,
innerrightmargin=0.7em, rightmargin=0.7em, innerleftmargin=1.4em, leftmargin=0.7em, middlelinewidth=.2em,
linecolor=commentColor, fontcolor=black, backgroundcolor=commentColor!15,
firstextra={\path let \p1=(P), \p2=(O) in (\x2,0.5 * \y1) node[commentsymbol] {\commentSymbol};},
secondextra={\path let \p1=(P), \p2=(O) in (\x2,0.5 * \y1) node[commentsymbol] {\commentSymbol};},
middleextra={\path let \p1=(P), \p2=(O) in (\x2,0.5 * \y1) node[commentsymbol] {\commentSymbol};},
singleextra={\path let \p1=(P), \p2=(O) in (\x2,0.5 * \y1) node[commentsymbol] {\commentSymbol};}
}

\mdfdefinestyle{question}{
hidealllines=true, leftline=true, skipabove=0pt, skipbelow=0pt, innertopmargin=0.7em, innerbottommargin=0.7em,
innerrightmargin=0.7em, rightmargin=0.7em, innerleftmargin=1.4em, leftmargin=0.7em, middlelinewidth=.2em,
linecolor=questionColor, fontcolor=black, backgroundcolor=questionColor!15,
firstextra={\path let \p1=(P), \p2=(O) in (\x2,0.5 * \y1) node[questionsymbol] {\questionSymbol};},
secondextra={\path let \p1=(P), \p2=(O) in (\x2,0.5 * \y1) node[questionsymbol] {\questionSymbol};},
middleextra={\path let \p1=(P), \p2=(O) in (\x2,0.5 * \y1) node[questionsymbol] {\questionSymbol};},
singleextra={\path let \p1=(P), \p2=(O) in (\x2,0.5 * \y1) node[questionsymbol] {\questionSymbol};}
}

\mdfdefinestyle{remark}{
hidealllines=true, leftline=true, skipabove=0pt, skipbelow=0pt, innertopmargin=0.7em, innerbottommargin=0.7em,
innerrightmargin=0.7em, rightmargin=0.7em, innerleftmargin=1.4em, leftmargin=0.7em, middlelinewidth=.2em,
linecolor=remarkColor, fontcolor=black, backgroundcolor=remarkColor!15,
firstextra={\path let \p1=(P), \p2=(O) in (\x2,0.5 * \y1) node[remarksymbol] {\remarkSymbol};},
secondextra={\path let \p1=(P), \p2=(O) in (\x2,0.5 * \y1) node[remarksymbol] {\remarkSymbol};},
middleextra={\path let \p1=(P), \p2=(O) in (\x2,0.5 * \y1) node[remarksymbol] {\remarkSymbol};},
singleextra={\path let \p1=(P), \p2=(O) in (\x2,0.5 * \y1) node[remarksymbol] {\remarkSymbol};}
}

\mdfdefinestyle{example}{
hidealllines=true, leftline=true, skipabove=0pt, skipbelow=0pt, innertopmargin=0.7em, innerbottommargin=0.7em,
innerrightmargin=0.7em, rightmargin=0.7em, innerleftmargin=1.4em, leftmargin=0.7em, middlelinewidth=.2em,
linecolor=exampleColor, fontcolor=black, backgroundcolor=exampleColor!15,
firstextra={\path let \p1=(P), \p2=(O) in (\x2,0.5 * \y1) node[examplesymbol] {\exampleSymbol};},
secondextra={\path let \p1=(P), \p2=(O) in (\x2,0.5 * \y1) node[examplesymbol] {\exampleSymbol};},
middleextra={\path let \p1=(P), \p2=(O) in (\x2,0.5 * \y1) node[examplesymbol] {\exampleSymbol};},
singleextra={\path let \p1=(P), \p2=(O) in (\x2,0.5 * \y1) node[examplesymbol] {\exampleSymbol};}
}

\mdfdefinestyle{proof}{
hidealllines=true, leftline=true, skipabove=0pt, skipbelow=0pt, innertopmargin=0.7em, innerbottommargin=0.7em,
innerrightmargin=0.7em, rightmargin=0.7em, innerleftmargin=1.4em, leftmargin=0.7em, middlelinewidth=.2em,
linecolor=proofColor, fontcolor=black, backgroundcolor=proofColor!15,
firstextra={\path let \p1=(P), \p2=(O) in (\x2,0.5 * \y1) node[proofsymbol] {\proofSymbol};},
secondextra={\path let \p1=(P), \p2=(O) in (\x2,0.5 * \y1) node[proofsymbol] {\proofSymbol};},
middleextra={\path let \p1=(P), \p2=(O) in (\x2,0.5 * \y1) node[proofsymbol] {\proofSymbol};},
singleextra={\path let \p1=(P), \p2=(O) in (\x2,0.5 * \y1) node[proofsymbol] {\proofSymbol};}
}

\newcommand{\kl}[2]{D_{\mathrm{KL}} \left[ \left. \left. #1 \right|\right| #2 \right] }
\DeclareMathOperator*{\argmin}{arg\,min}
\newcommand{\indep}{\perp \! \! \! \perp}

\makeatletter
\newcommand*\bigcdot{\mathpalette\bigcdot@{.5}}
\newcommand*\bigcdot@[2]{\mathbin{\vcenter{\hbox{\scalebox{#2}{$\m@th#1\bullet$}}}}}
\makeatother

\begin{document}

\title{Reframing the Expected Free Energy:\\\small{Four Formulations and a Unification.}}

\author{\name Théophile Champion \email tmac3@kent.ac.uk \\
       \addr University of Kent, School of Computing\\
       Canterbury CT2 7NZ, United Kingdom
       \AND
       \name Howard Bowman \email H.Bowman@kent.ac.uk \\
       \addr University of Birmingham, School of Psychology and School of Computer Science,\\
       Birmingham B15 2TT, United Kingdom\\
       University of Kent, School of Computing\\
       Canterbury CT2 7NZ, United Kingdom\\
       University College London, Wellcome Centre for Human Neuroimaging (honorary)\\
       London WC1N 3AR, United Kingdom
       \AND
       \name Dimitrije Marković \email dimitrije.markovic@tu-dresden.de \\
       \addr Technische Universität Dresden, Department of Psychology \\
	   Dresden 01069, Germany
	   \AND
       \name Marek Grze\'s \email m.grzes@kent.ac.uk \\
       \addr University of Kent, School of Computing\\
       Canterbury CT2 7NZ, United Kingdom
       }
       
\editor{\textbf{TO BE FILLED}} 

\maketitle

\begin{abstract} 
Active inference is a leading theory of perception, learning and decision making, which can be applied to neuroscience, robotics, psychology, and machine learning. Active inference is based on the expected free energy, which is mostly justified by the intuitive plausibility of its formulations, e.g., the risk plus ambiguity and information gain / pragmatic value formulations. This paper seek to formalize the problem of deriving these formulations from a single root expected free energy definition, i.e., the unification problem. Then, we study two settings, each one having its own root expected free energy definition. In the first setting, no justification for the expected free energy has been proposed to date, but all the formulations can be recovered from it. However, in this setting, the agent cannot have arbitrary prior preferences over observations. Indeed, only a limited class of prior preferences over observations is compatible with the likelihood mapping of the generative model. In the second setting, a justification of the root expected free energy definition is known, but this setting only accounts for two formulations, i.e., the risk over states plus ambiguity and entropy plus expected energy formulations.
\end{abstract}
\vspace{0.5cm}
\begin{keywords}
Active Inference, Expected Free Energy, Unification Problem
\end{keywords}

\section{Introduction}

Active inference \citep{FRISTON2016862,bayes_surprise,curiosity,dopamine,DeepAIwithMCMC,sancaktar2020endtoend,catal2020learning,CULLEN2018809,cart_pole} is a framework for decision-making under uncertainty, in which the agent is equipped with a (generative) model that encodes the environment dynamics, and a variational posterior approximating the true posterior over latent variables. The variational posterior is computed by minimizing a function called the variational free energy (VFE), also known as the negative evidence lower bound in machine learning \citep{VI_TUTO,VMP_TUTO}. While the variational posterior defines the most likely state of the environment, it does not indicate which action should be selected. Instead, the agent aims to reach a set of preferred states or observations, by minimizing the expected free energy (EFE).

While the variational free energy has a clear root definition that all other formulations are derived from, the literature does not clearly identify such a root definition for the expected free energy, leaving the question of antecedence amongst its many formulations as moot. 

The EFE is a function defining the cost of performing a particular policy as a trade-off between exploration and exploitation, e.g., the goal is to maximize pragmatic value (reward), while also maximizing information gain. The pragmatic value relies on the prior preferences of the agent, which specify the preferred states or observations, and provides the agent with its goal-directed behaviour.

In the active inference literature, the prior preferences over observation are usually denoted $P(\overline{o} ; C)$, where $C$ is a vector of parameters, i.e., there is no distribution over $C$. Thus, the prior preferences over observation can also be written $P(\overline{o})$. Importantly, the prior preferences are denoted using the letter $P$, suggesting that these preferences are part of the generative model. However, $P(\overline{o})$ in the generative model refers to the marginal likelihood. Therefore, the symbol $P(\overline{o})$ has a dual meaning, i.e., it refers to both the prior preferences and the marginal likelihood. This dual meaning may not seem problematic at first, but introduces hidden inconsistencies. For example, in most cases, the prior preferences will be constant, i.e., not updating during a run. However, in general, the prior over future state $P(\overline{s})$ will change as time passes and the agent gathers new observations, thus the marginal likelihood will change, suggesting an inconsistency resulting from the conflation of the two interpretations. More formally, when using Bayes theorem:
$$P(\overline{s} | \overline{o}) = \frac{P(\overline{o} | \overline{s})P(\overline{s})}{P(\overline{o})},$$
the denominator corresponds to the joint distribution $P(\overline{o}, \overline{s})$ marginalized over the states $\overline{s}$. However, since the prior preferences are also denoted $P(\overline{o})$, the denominator could be interpreted as being the prior preferences, but in most cases: $P(\overline{o}) \neq P(\overline{o} ; C)$. Importantly, the dual meaning problem can also emerge if the joint distribution $P(\overline{o}, \overline{s})$ is marginalized using the sum-rule of probability or split using the product rule, i.e.,
$$P(\overline{o}) = \sum_{\overline{s}} P(\overline{o}, \overline{s}) \quad \text{or} \quad P(\overline{o}, \overline{s}) = P(\overline{s} | \overline{o})P(\overline{o}).$$
\noindent To solve the dual meaning problem, the prior preferences are sometime considered as being part of a target distribution. However, in this paper, we show that this assumption restricts the class of valid prior preferences, and leads to a definition of the expected free energy that is not currently justified.
In the following sections, we explore two possible interpretations of \citet{parr2022active}, and explain their limitation. Appendix B and C provide a description of the properties used throughout this paper.

\section{Generative model}

In active inference, the agent is equipped with a (generative) model of its environment that spans all time steps until the present time $t$. This model is composed of (a) hidden states $s_{0:t}$ representing states of the environment that the agent does not directly observe, (b) observations $o_{0:t}$, which represent measurements made by the agent, and (c) actions $a_{0:t-1}$ that the agent performed in the environment. For the sake of conciseness, $s_{0:t}$, $o_{0:t}$, and $a_{0:t-1}$ will be denoted $\underline{s}$, $\underline{o}$, and $\underline{a}$, repectively. Moreover, in this paper, we assume that observations depend on states, and each state depends on the state and action at the previous time step. Formally, this setting is called a Partially Observable Markov Decision Process (POMDP), and the model definition is as follows:
$$P(\underline{o}, \underline{s} | \underline{a}) \delequal P(s_0) \prod_{\tau = 0}^{t} P(o_\tau | s_\tau) \prod_{\tau = 1}^{t} P(s_\tau | s_{\tau - 1}, a_{\tau - 1}).$$

\section{Variational distribution}

The generative model described in the previous section encodes prior beliefs about the environment dynamics. However, when making measurements of key quantities, e.g., $\underline{o}$, the agent needs to compute posterior beliefs about the states, e.g., $P(\underline{s} | \underline{o}, \underline{a})$. These posterior beliefs encode the new beliefs of the agent when taking into consideration the new observations. Unfortunately, computing the true posterior can either be analytically intractable or simply too computationally expensive. Therefore, the true posterior is generally approximated by a variational distribution $Q(\underline{s} | \underline{a})$:

$$\underbrace{Q(\underline{s} | \underline{a})}_{\text{variational\,\, posterior}} \approx \quad \underbrace{P(\underline{s} | \underline{o}, \underline{a})}_{\text{true\,\, posterior}} \quad \propto \quad \underbrace{P(\underline{o}, \underline{s} | \underline{a})}_{\text{generative\,\, model}}$$

\noindent In active inference, the variational posterior 1) factorises over time steps, i.e., a temporal mean-field approximation, but 2) all states still depend on the policy $\underline{a}$. These two assumptions lead to the following definition of the variational distribution:
$$Q(\underline{s} | \underline{a}) \delequal \prod_{\tau = 0}^{t} Q(s_\tau | \underline{a})$$

\section{Variational inference and the variational free energy}

To sum up, the agent is provided with a generative model $P(\underline{o}, \underline{s} | \underline{a})$ and a variational distribution $Q(\underline{s} | \underline{a})$. Given some measurements $\underline{o}$, the variational distribution needs to approximate the true posterior $P(\underline{s} | \underline{o}, \underline{a})$. This can be formally expressed as minimising the Kullback-Leibler divergence between the approximate and true posteriors:

\begin{align*}
Q^*(\underline{s} | \underline{a}) &= \argmin_{Q(\underline{s} | \underline{a})} \kl{Q(\underline{s} | \underline{a})}{P(\underline{s} | \underline{o}, \underline{a})}\phantom{\Bigg[}
\end{align*}

\noindent Minimising this KL-divergence and minimising the variational free energy (VFE) is equivalent (see proof below). Intuitively, the VFE trades-off accuracy, i.e., how well are the observations predicted, and complexity, i.e., how far the posterior is from the prior. More formally, the VFE is defined as follows:

\begin{align*}
\mathcal{F}\Big[Q(\underline{s} | \underline{a}), \underline{a}, \underline{o}\Big] &= \underbrace{\kl{Q(\underline{s} | \underline{a})}{P(\underline{s}| \underline{a})}}_{\text{complexity}} - \underbrace{\mathbb{E}_{Q(\underline{s} | \underline{a})}[\ln P(\underline{o} | \underline{s})]}_{\text{accuracy}}\phantom{\Bigg[}
\end{align*}

\begin{mdframed}[style=proof]
The derivation of the variational free energy proceeds as follows:
\begin{align*}
Q^*(\underline{s} | \underline{a}) &= \argmin_{Q(\underline{s} | \underline{a})} \kl{Q(\underline{s} | \underline{a})}{P(\underline{s} | \underline{o}, \underline{a})}\phantom{\Bigg[}\\
&= \argmin_{Q(\underline{s} | \underline{a})} \mathbb{E}_{Q(\underline{s} | \underline{a})}\Bigg[\ln Q(\underline{s} | \underline{a}) - \ln \frac{P(\underline{o} | \underline{s})P(\underline{s}| \underline{a})}{P(\underline{o} | \underline{a})}\Bigg]\\
&= \argmin_{Q(\underline{s} | \underline{a})} \mathbb{E}_{Q(\underline{s} | \underline{a})}[\ln Q(\underline{s} | \underline{a}) - \ln P(\underline{o} | \underline{s})P(\underline{s}| \underline{a})] + \underbrace{\ln P(\underline{o} | \underline{a})}_{\text{constant}}\phantom{\Bigg[}\\
&= \argmin_{Q(\underline{s} | \underline{a})} \underbrace{\kl{Q(\underline{s} | \underline{a})}{P(\underline{s}| \underline{a})}}_{\text{complexity}} - \underbrace{\mathbb{E}_{Q(\underline{s} | \underline{a})}[\ln P(\underline{o} | \underline{s})]}_{\text{accuracy}}\phantom{\Bigg[}\\
&= \argmin_{Q(\underline{s} | \underline{a})} \underbrace{\mathcal{F}\Big[Q(\underline{s} | \underline{a}), \underline{a}, \underline{o}\Big]}_{\text{variational\,\,free\,\,energy}}, \phantom{\Bigg[}
\end{align*}
where the d-separation criterion \citep{koller2009probabilistic} (which simplifies $P(\underline{o} | \underline{s}, \underline{a})$ to $P(\underline{o} | \underline{s})$ in the second line, given the independence of $\underline{o}$ and $\underline{a}$ given $\underline{s}$) and Bayes theorem have been used, along with the linearity of expectation and the log-property.
\end{mdframed}
\vspace{0.5cm}
\begin{mdframed}[style=comment]
In this paper, we focus on planning, therefore we do not explain how the optimal variational distribution $Q^*(\underline{s} | \underline{a})$ is computed. The interested reader is referred to \citep{CHAMPION2022295,parr2019neuronal, graphical_brain,winn2005variational} for an approximate scheme based on variational message passing, and \citep{BTAI_3MF,kschischang2001factor} for an exact scheme based on the sum-product algorithm.
\end{mdframed}

\section{Planning and the expected free energy} \label{sec:planning}

After performing inference, the agent has at its disposal posterior beliefs encoded by the optimal variational distribution $Q^*(\underline{s} | \underline{a})$. At this point, the agent needs to choose the next action to perform in the environment. In active inference, the cost of a policy is given by the expected free energy (EFE), thus the goal of planning is to identify the policy with the smallest EFE.
\newline\\
\noindent Unfortunately, given a time horizon of planning $h$, the number of policies is $|\mathcal{A}|^{h - t}$, where  $\mathcal{A}$ is the set of all actions available to the agent, and $|\mathcal{A}|$ is the cardinality of this set. As the number of policies grows exponentially with the time horizon, computing the EFE of all policies requires exponential time. Therefore, it is important to search the space of policies efficiently. For example, one can use Monte-Carlo tree search \citep{BTAI_empirical,BTAI_BF}, which maintains a balance between exploiting policies with low EFE and exploring rarely visited action sequences. Another solution would be to use sophisticated inference \citep{sophisticated_inference}, which implements a tree search over actions and outcomes in the future, by using a recursive form of expected free energy.
\newline\\
\noindent In this paper, we postulate that the expected free energy is based on two distributions: the forecast $F(\overline{o}, \overline{s}| \overline{a})$ and target $T(\overline{o}, \overline{s}| \overline{a})$ distributions, where the future observations, states, and actions are denoted $\overline{o} = o_{t+1:h}$, $\overline{s} = s_{t+1:h}$ and $\overline{a} = a_{t:h-1}$, respectively. The forecast distribution predicts the future according to the agent's best beliefs about the current states of the environment, and its generative model. In contrast, the target distribution encodes the states and observations the agent wants to reach. A general formulation of the target is given here, allowing it to change with policies. Although, in most cases, it will not change.

\vspace{0.5cm}
\begin{mdframed}[style=question]
	Why is the forecast distribution introduced?
\end{mdframed}
\vspace{0.5cm}
\begin{mdframed}[style=answer]
	In the active inference literature, it is frequent to see Bayes theorem being used between factors of the generative model $P(\underline{o}, \underline{s}| \underline{a})$ and the variational distribution $Q(\underline{s}| \underline{a})$, or even to see factors from the generative model $P(x|y)$ being replaced by their variational counterpart $Q(x|y)$. However, Bayes theorem is a corollary of the product rule of probability, i.e.,
	$$P(x, y) = P(x|y)P(y) = P(y|x)P(x) \Leftrightarrow P(y|x) = \frac{P(x|y)P(y)}{P(x)}.$$
	Thus, technically, Bayes theorem cannot be used between factors of two different distributions. To be really explicit, equality between $P(y|x)$ and the right-hand-side would not hold if, for example, $P(y)$ were replaced by any distribution different to $P(y)$, and similarly for $P(x|y)P(y)$ and $P(x)$. Importantly, it is straightforward to see that the generative model and the variational distribution are two different distributions. The easiest way to see this, is to realise that these two distributions do not even share the same domain, i.e., the generative model is a distribution over states $\underline{s}$ and observations $\underline{o}$, while the variational distribution is a distribution over states $\underline{s}$ only. One might consider observations to be implicitly present, but in this respect, they are a ground term, with a particular value, rather than random variables with a distribution over possible values. Another intuitive argument is that the generative model encodes the prior beliefs of the agent, then upon receiving new data, the agent computes its (approximate) posterior beliefs. If the prior was equal to the posterior, there will be no point in performing inference in the first place. Since the generative model and the variational distribution are two different distributions, one should be careful when replacing factors from one distribution by factors from the other. It is for this reason that the forecast distribution is introduced. Put simply, the forecast distribution provides a bridge between factors of the variational posterior and those of the generative model.
\end{mdframed}

\subsection{The unification problem}

In this section, we formalise the problem of deriving the four EFE formulations, which can be found in the literature (see below), from the EFE definition, i.e., \textit{the unification problem}. Specifically, the unification problem is a 4-tuple $\mathcal{P} = \langle F, T, \mathcal{G}_{rt}, \mathcal{C} \rangle$, where $F$ is the forecast distribution, $T$ is the target distribution, $\mathcal{G}_{rt}$ is the definition of the expected free energy, and $\mathcal{C} = \{\mathcal{C}_{RSA}, \mathcal{C}_{ROA}, \mathcal{C}_{IGPV}, \mathcal{C}_{3E}\}$ is a set containing the four formulations of the expected free energy. Solving $\mathcal{P}$ consists of finding a definition of $F$, $T$, and $\mathcal{G}_{rt}$ such that it is possible to derive $\mathcal{C}_X$ from $\mathcal{G}_{rt}$ for all $\mathcal{C}_X \in \mathcal{C}$.
\newline\\
\noindent We now define the four formulations of the expected free energy, which are based upon the definitions in \cite{parr2022active}. The formulation for the risk over states and ambiguity is as follows:
\begin{align*}
\mathcal{C}_{RSA}(\overline{a}) &\delequal \underbrace{\kl{F(\overline{s} | \overline{a})}{T(\overline{s} | \overline{a})}}_{\text{risk over states}} + \underbrace{\mathbb{E}_{F(\overline{s} | \overline{a})} \big[H[F(\overline{o} | \overline{s})]\big]}_{\text{ambiguity}}.
\end{align*}
Importantly, the risk over states is the KL-divergence between the predictive posterior over states $F(\overline{s} | \overline{a})$ and the prior preferences over states $T(\overline{s} | \overline{a})$, and the ambiguity is the expected entropy of the likelihood mapping according to the generative model. The risk over states pushes the predictive posterior towards the prior preferences, while the ambiguity encourages the agent to visit states producing a low entropy distribution over observations, i.e., if we arrive at a state, we know which observation(s) to expect. The formulation for the risk over observations and ambiguity is as follows:
\begin{align*}
\mathcal{C}_{ROA}(\overline{a}) &\delequal \underbrace{\kl{F(\overline{o} | \overline{a})}{T(\overline{o} | \overline{a})}}_{\text{risk over observations}} + \underbrace{\mathbb{E}_{F(\overline{s} | \overline{a})} \big[H[F(\overline{o} | \overline{s})]\big]}_{\text{ambiguity}}.
\end{align*}
The ambiguity term is identical, and the risk over observations is a KL-divergence, which pushes the predictive posterior over observations $F(\overline{o} | \overline{a})$ to be as close as possible to the prior preferences over observations $T(\overline{o} | \overline{a})$. The formulation for the information gain and pragmatic value is as follows:
\begin{align*}
\mathcal{C}_{IGPV}(\overline{a}) &\delequal - \underbrace{\mathbb{E}_{F(\overline{o} | \overline{a})} \big[\kl{F(\overline{s} | \overline{o}, \overline{a})}{F(\overline{s} | \overline{a})\big]}}_{\text{information gain}} - \underbrace{\mathbb{E}_{F(\overline{o} | \overline{a})} [\ln T(\overline{o} | \overline{a})]}_{\text{pragmatic value}}. \phantom{\sum}
\end{align*}
Importantly, the information gain is a KL-divergence that relies only on factors from the forecast distribution. This prevents degenerate behaviours were the agent stops exploring its environment, i.e., information loss \citep{deconstructing_DAI}. In addition, the pragmatic value is based on the preferred observations $T(\overline{o} | \overline{a})$, which provides the agent with its goal directed behaviour. Finally, the expected energy and entropy formulation is as follows:
\begin{align*}
\mathcal{C}_{3E}(\overline{a}) &\delequal - \underbrace{H[F(\overline{s} | \overline{a})]}_{\text{entropy}} - \underbrace{\mathbb{E}_{F(\overline{o}, \overline{s} | \overline{a})} [\ln T(\overline{o}, \overline{s} | \overline{a})]}_{\text{expected energy}} \phantom{\sum}
\end{align*}
The entropy term ensures that a good policy is one which keeps our options open by allowing us to reach a wide variety of states, as implied by Jaynes' theory of maximum entropy \citep{jaynes1957information, jaynes1957information2}. Additionally, as shown in the proof below, the expected energy encourages the agent to reach its preferred states, while also pushing the agent to select states for which the associated distribution over observations has low entropy, i.e., given a state, we know which observation(s) to expect.

\vspace{0.5cm}
\begin{mdframed}[style=proof]
Starting with the negative expected energy, one can use the product rule, the log-property and the linearity of expectation to get:
\begin{align*}
- \underbrace{\mathbb{E}_{F(\overline{o}, \overline{s} | \overline{a})} [\ln T(\overline{o}, \overline{s} | \overline{a})]}_{\text{expected energy}} = - \mathbb{E}_{F(\overline{o}, \overline{s}| \overline{a})} [\ln T(\overline{o} | \overline{s}, \overline{a})] - \mathbb{E}_{F(\overline{s}| \overline{a})} [\ln T(\overline{s} | \overline{a})].
\end{align*}
Then, assuming that the forecast distribution is a partially observable Markov decision process, and that the likelihood of the forecast and target distributions are the same, we obtain:
\begin{align*}
- \underbrace{\mathbb{E}_{F(\overline{o}, \overline{s} | \overline{a})} [\ln T(\overline{o}, \overline{s} | \overline{a})]}_{\text{expected energy}} &= - \mathbb{E}_{F(\overline{o}, \overline{s}| \overline{a})} [\ln F(\overline{o} | \overline{s}, \overline{a})] - \mathbb{E}_{F(\overline{s}| \overline{a})} [\ln T(\overline{s} | \overline{a})] \\
&= - \mathbb{E}_{F(\overline{o}, \overline{s}| \overline{a})} [\ln F(\overline{o} | \overline{s})] - \mathbb{E}_{F(\overline{s}| \overline{a})} [\ln T(\overline{s} | \overline{a})]
\end{align*}
where, $F(\overline{o}|\overline{s}, \overline{a}) = F(\overline{o}|\overline{s})$, due to d-separation, observing that $\overline{o}$ and $\overline{a}$ are conditionally independent given $\overline{s}$. Lastly, using this same property again after applying the product rule and then using the definition of entropy gives the final result:
\begin{align*}
- \underbrace{\mathbb{E}_{F(\overline{o}, \overline{s} | \overline{a})} [\ln T(\overline{o}, \overline{s} | \overline{a})]}_{\text{expected energy}} = \underbrace{\mathbb{E}_{F(\overline{s}| \overline{a})} [H[F(\overline{o} | \overline{s})]]}_{\text{ambiguity}} - \underbrace{\mathbb{E}_{F(\overline{s}| \overline{a})} [\ln T(\overline{s} | \overline{a})]}_{\text{pragmatic value}},
\end{align*}
This equation shows that maximising expected energy means selecting states that minimise the ambiguity of the likelihood while maximising the pragmatic value of states.
\end{mdframed}

\subsection{Forecast distribution}

As previously mentioned, the forecast distribution predicts the future according to the agent's best beliefs about the current states of the environment, and its generative model. More formally, the forecast distribution factorizes as follows:
$$F(\overline{o}, \overline{s} | \overline{a}) \delequal F(s_{t+1} | a_t) \prod_{\tau = t + 1}^{h} F(o_\tau | s_\tau) \prod_{\tau = t + 2}^{h} F(s_\tau | s_{\tau-1}, a_{\tau-1}).$$
Additionally, we make three assumptions to define the factors of the forecast distribution. These assumptions define the forecast distribution in terms of factors from the generative model and the variational distribution. But importantly, these definitions are made explicit here, leaving no uncertainty about these relationships amongst probability distributions. First, we assume that the likelihood in the future $F(o_\tau | s_\tau)$ is the same as in the past $P(o_\tau | s_\tau)$, i.e., the likelihood of the forecast distribution is the same as the likelihood  of the generative model. Second, the temporal transition in the future $F(s_\tau | s_{\tau - 1}, a_{\tau - 1})$ is the same as in the past $P(s_\tau | s_{\tau - 1}, a_{\tau - 1})$, i.e., the temporal transition of the forecast distribution is the same as the temporal transition of the generative model. Third, the agent's best prior over the current state $F(s_{t})$ is given by the optimal variational posterior $Q^*(s_{t} | \underline{a})$. More formally:
\begin{align*}
	F(o_\tau | s_\tau) &= P(o_\tau | s_\tau)\\
	F(s_\tau | s_{\tau - 1}, a_{\tau - 1}) &= P(s_\tau | s_{\tau - 1}, a_{\tau - 1}) \\
	F(s_t) &= Q^*(s_{t} | \underline{a})
\end{align*}
Using the above assumptions, we obtain $F(s_{t+1}|a_t)$ by taking the expectation of the temporal transition w.r.t. the prior over states at time $t$. For generality, we define this as an integral, but this could be specified to a summation in the discrete case:
\begin{align*}
	F(s_{t+1}|a_t) &= \int_{s_t} F(s_{t+1} | s_t, a_t) F(s_{t})\,\, ds_t = \int_{s_t} P(s_{t+1} | s_t, a_t) Q^*(s_{t} | \underline{a})\,\, ds_t
\end{align*}

\subsection{Target distribution}

The second distribution of interest is the target distribution, which encodes the states and observations that the agent wants to reach. In the following section, we define the target distribution as follows:
$$T(\overline{o}, \overline{s} | \overline{a}) \delequal \prod_{\tau = t + 1}^{h} T(o_\tau | s_\tau)T(s_\tau | \overline{a}),$$
where $T(o_\tau | s_\tau) = P(o_\tau | s_\tau)$ and $T(s_\tau | \overline{a}) = \text{Cat}(s_\tau; \bm{C}_s)$. Note, $\bm{C}_s$ is the matrix known as the $\bm{C}$ matrix in the active inference literature, where the prior preferences are defined w.r.t. states. Also, the target's likelihood is equal to $P(o_\tau | s_\tau)$, which means that given any state, the agent wants to reach the observations that will arise naturally according to the likelihood of the generative model, i.e., $P(o_\tau | s_\tau)$. If the target changes with policy $\overline{a}$, this would have to be built into the $\bm{C}_s$ matrix, but in most cases, the target distribution would be fixed.

\subsection{Solving the unification problem} \label{ssec:solving_the_unification_problem}

With the forecast and target distributions laid out, we now focus on the unification problem. We will explore whether any of the EFE formulations could serve as a root definition from which all other formulations could be derived. First, we define the root expected free energy as the risk over observations plus ambiguity:
\begin{align}
	\mathcal{G}_{rt}(\overline{a}) &\delequal \underbrace{\kl{F(\overline{o} | \overline{a})}{T(\overline{o} | \overline{a})}}_{\text{risk over observations}} + \underbrace{\mathbb{E}_{F(\overline{s} | \overline{a})} \big[H[F(\overline{o} | \overline{s})]\big]}_{\text{ambiguity}} = \mathcal{C}_{ROA}(\overline{a}). \label{eq:efe_definition}
\end{align}

\subsubsection{The information gain / pragmatic value formulation}

\noindent In this section, we demonstrate that the information gain / pragmatic value formulation can be recovered from the $\mathcal{C}_{ROA}(\overline{a})$-as-root expected free energy definition. The derivation relies on the following equality:
\begin{align}
	\frac{F(\overline{s} | \overline{a})}{F(\overline{s} | \overline{o}, \overline{a})} = \frac{F(\overline{o} | \overline{a})}{F(\overline{o} | \overline{s})}, \label{eq:pomdp_bayes_theorem}
\end{align}
which holds because the forecast distribution is a partially observable Markov decision process.
\vspace{0.5cm}
\begin{mdframed}[style=proof]
	We start by re-arranging Bayes theorem as follows:
	$$F(\overline{s} | \overline{o}, \overline{a}) = \frac{F(\overline{o} | \overline{s}, \overline{a})F(\overline{s} | \overline{a})}{F(\overline{o} | \overline{a})} \Leftrightarrow \frac{F(\overline{s} | \overline{a})}{F(\overline{s} | \overline{o}, \overline{a})} = \frac{F(\overline{o} | \overline{a})}{F(\overline{o} | \overline{s}, \overline{a})}.$$
	Then, in a partially observable Markov decision process, $\overline{o} \indep \overline{a} \,\, | \,\, \overline{s}$, (i.e., the Markov property ensures that observation sequences are conditionally independent of policies, if the sequence of states is known), thus $F(\overline{o} | \overline{s}, \overline{a}) = F(\overline{o} | \overline{s})$ and:
	$$\frac{F(\overline{s} | \overline{a})}{F(\overline{s} | \overline{o}, \overline{a})} = \frac{F(\overline{o} | \overline{a})}{F(\overline{o} | \overline{s})}.$$
\end{mdframed}
\vspace{0.5cm}
Importantly, by starting with the definition of $\mathcal{C}_{IGPV}(\overline{a})$ and using \eqref{eq:pomdp_bayes_theorem}, one can show that:
\begin{align}
	\mathcal{G}_{rt}(\overline{a}) &= - \underbrace{\mathbb{E}_{F(\overline{o} | \overline{a})} [\kl{F(\overline{s} | \overline{o}, \overline{a})}{F(\overline{s} | \overline{a})} ]}_{\text{information gain}}  - \underbrace{\mathbb{E}_{F(\overline{o} | \overline{a})} [\ln T(\overline{o} | \overline{a})]}_{\text{pragmatic value}} = \mathcal{C}_{IGPV}(\overline{a}). \phantom{\sum} \label{eq:info_gain_prag_value_setting_1}
\end{align}
\vspace{0.15cm}
\begin{mdframed}[style=proof]
	Starting with the definition of $\mathcal{C}_{IGPV}(\overline{a})$:
	\begin{align*}
		\mathcal{C}_{IGPV}(\overline{a}) &= - \underbrace{\mathbb{E}_{F(\overline{o} | \overline{a})} [\kl{F(\overline{s} | \overline{o}, \overline{a})}{F(\overline{s} | \overline{a})} ]}_{\text{information gain}}  - \underbrace{\mathbb{E}_{F(\overline{o} | \overline{a})} [\ln T(\overline{o} | \overline{a})]}_{\text{pragmatic value}}, \phantom{\sum}
	\end{align*}
	and using the KL-divergence definition, and that $F(\overline{o}, \overline{s} | \overline{a}) = F(\overline{s} | \overline{o}, \overline{a})F(\overline{o} | \overline{a})$, by the product rule, we obtain:
	\begin{align*}
		\mathcal{C}_{IGPV}(\overline{a}) &= - \mathbb{E}_{F(\overline{o}, \overline{s} | \overline{a})} [\ln F(\overline{s} | \overline{o}, \overline{a}) - \ln F(\overline{s} | \overline{a}) ]  - \mathbb{E}_{F(\overline{o} | \overline{a})} [\ln T(\overline{o} | \overline{a})] \phantom{\sum}
	\end{align*}
	Then, by using the log-properties and \eqref{eq:pomdp_bayes_theorem} to replace $\frac{F(\overline{s} | \overline{o}, \overline{a})}{F(\overline{s} | \overline{a})}$ by $\frac{F(\overline{o} | \overline{s})}{F(\overline{o} | \overline{a})}$, we get:
	\begin{align*}
		\mathcal{C}_{IGPV}(\overline{a}) &= - \mathbb{E}_{F(\overline{o}, \overline{s} | \overline{a})} [\ln F(\overline{o} | \overline{s}) - \ln F(\overline{o} | \overline{a}) ]  - \mathbb{E}_{F(\overline{o} | \overline{a})} [\ln T(\overline{o} | \overline{a})] \phantom{\sum}
	\end{align*}
	Next, the linearity of expectation can be applied to re-arrange the expression as follows:
	\begin{align*}
		\mathcal{C}_{IGPV}(\overline{a}) &= - \mathbb{E}_{F(\overline{o}, \overline{s} | \overline{a})} [\ln F(\overline{o} | \overline{s}) ]  + \mathbb{E}_{F(\overline{o} | \overline{a})} [\ln F(\overline{o} | \overline{a}) - \ln T(\overline{o} | \overline{a})] \phantom{\sum}
	\end{align*}
	Lastly, recognizing the entropy and KL-divergence definitions leads to the final results:
	\begin{align*}
		\mathcal{C}_{IGPV}(\overline{a}) = \underbrace{\kl{F(\overline{o} | \overline{a})}{T(\overline{o} | \overline{a})}}_{\text{risk over observations}} + \underbrace{\mathbb{E}_{F(\overline{s} | \overline{a})} \big[H[F(\overline{o} | \overline{s})]\big]}_{\text{ambiguity}} = \mathcal{G}_{rt}(\overline{a}).
	\end{align*}
\end{mdframed}

\subsubsection{The risk over states vs ambiguity formulation} \label{subsubsection:RSA_formulation_second_setting}

\noindent In this section, we demonstrate that the risk over states plus ambiguity is an upper bound of the expected free energy. Restarting from the EFE definition, one can show that: 
\begin{align*}
	\mathcal{G}_{rt}(\overline{a}) &\leq \kl{F(\overline{o}, \overline{s} | \overline{a})}{T(\overline{o}, \overline{s} | \overline{a})} + \mathbb{E}_{F(\overline{s} | \overline{a})} \big[H[F(\overline{o} | \overline{s})]\big]. \phantom{\sum}
\end{align*}

\begin{mdframed}[style=proof]
	We follow the proof provided in Appendix B of \cite{parr2022active}, but using our notation and going from the end of the proof to the beginning. Restarting from the EFE definition:
	\begin{align*}
		\mathcal{G}_{rt}(\overline{a}) &= \underbrace{\kl{F(\overline{o} | \overline{a})}{T(\overline{o} | \overline{a})}}_{\text{risk over observations}} + \underbrace{\mathbb{E}_{F(\overline{s} | \overline{a})} \big[H[F(\overline{o} | \overline{s})]\big]}_{\text{ambiguity}}, \phantom{\sum}
	\end{align*}
	we obtain an upper bound on the EFE by adding the following bound, which is the expectation of a KL-divergence and cannot be negative:
	\begin{align*}
		\mathcal{G}_{rt}(\overline{a}) &\leq \underbrace{\kl{F(\overline{o} | \overline{a})}{T(\overline{o} | \overline{a})}}_{\text{risk over observations}} + \, \mathbb{E}_{F(\overline{s} | \overline{a})} \big[H[F(\overline{o} | \overline{s})]\big] + \, \underbrace{\mathbb{E}_{F(\overline{o} | \overline{a})} \big[\kl{F(\overline{s} | \overline{o}, \overline{a})}{T(\overline{s} | \overline{o}, \overline{a})}\big]}_{\text{bound}}. \phantom{\sum}
	\end{align*}
	Next, using the linearity of expectation and the log-property, the bound can be merged to the risk over observations:
	\begin{align}
		\mathcal{G}_{rt}(\overline{a}) &\leq \kl{F(\overline{o}, \overline{s} | \overline{a})}{T(\overline{o}, \overline{s} | \overline{a})} + \mathbb{E}_{F(\overline{s} | \overline{a})} \big[H[F(\overline{o} | \overline{s})]\big]. \phantom{\sum} \label{eq:efe_joint_setting_2_from_book}
	\end{align}
\end{mdframed}
\vspace{0.5cm}
Additionally, if one assumes that $T(\overline{o} | \overline{s}) = F(\overline{o} | \overline{s})$, then restarting from Equation \ref{eq:efe_joint_setting_2_from_book}, one can show that the risk over states plus ambiguity is an upper bound of the expected free energy, i.e.,
\begin{align}
	\mathcal{G}_{rt}(\overline{a}) &\leq \kl{F(\overline{s} | \overline{a})}{T(\overline{s} | \overline{a})} + \mathbb{E}_{F(\overline{s} | \overline{a})} \big[H[F(\overline{o} | \overline{s})]\big] = \mathcal{C}_{RSA}(\overline{a}). \label{eq:risk_over_states_plus_ambiguity_in_setting_2} \phantom{\sum}
\end{align}

\begin{mdframed}[style=proof]
	Once again, we keep following the proof presented in Appendix B of \cite{parr2022active} backward. Let us restart from Equation \ref{eq:efe_joint_setting_2_from_book}:
	\begin{align*}
		\mathcal{G}_{rt}(\overline{a}) &\leq \kl{F(\overline{o}, \overline{s} | \overline{a})}{T(\overline{o}, \overline{s} | \overline{a})} + \mathbb{E}_{F(\overline{s} | \overline{a})} \big[H[F(\overline{o} | \overline{s})]\big]. \phantom{\sum} 
	\end{align*}
	Then, using the definition of the KL-divergence, the linearity of expectation and the log-property, we can split the KL-divergence as follows:
	\begin{align*}
		\mathcal{G}_{rt}(\overline{a}) &\leq \kl{F(\overline{s} | \overline{a})}{T(\overline{s} | \overline{a})} + \mathbb{E}_{F(\overline{s} | \overline{a})} \big[H[F(\overline{o} | \overline{s})]\big]  \phantom{\sum} \\
		& \quad + \mathbb{E}_{F(\overline{o}, \overline{s} | \overline{a})} [\ln F(\overline{o} | \overline{s})] - \mathbb{E}_{F(\overline{o}, \overline{s} | \overline{a})} [\ln T(\overline{o} | \overline{s})]. \phantom{\sum}
	\end{align*}
	Next, using our assumption that $T(\overline{o} | \overline{s}) = F(\overline{o} | \overline{s})$, we can get:
	\begin{align*}
		\mathcal{G}_{rt}(\overline{a}) &\leq \kl{F(\overline{s} | \overline{a})}{T(\overline{s} | \overline{a})} + \mathbb{E}_{F(\overline{s} | \overline{a})} \big[H[F(\overline{o} | \overline{s})]\big]  \phantom{\sum} \\
		& \quad + \underbrace{\mathbb{E}_{F(\overline{o}, \overline{s} | \overline{a})} [\ln F(\overline{o} | \overline{s})] - \mathbb{E}_{F(\overline{o}, \overline{s} | \overline{a})} [\ln F(\overline{o} | \overline{s})]}_{\text{=0}}, \phantom{\sum}
	\end{align*}
	which simplifies to:
	\begin{align*}
		\mathcal{G}_{rt}(\overline{a}) &\leq \underbrace{\kl{F(\overline{s} | \overline{a})}{T(\overline{s} | \overline{a})}}_{\text{risk over states}} + \underbrace{\mathbb{E}_{F(\overline{s} | \overline{a})} \big[H[F(\overline{o} | \overline{s})]\big]}_{\text{ambiguity}} = \mathcal{C}_{RSA}(\overline{a}). \phantom{\sum}
	\end{align*}
\end{mdframed}
\vspace{0.5cm}
\noindent Importantly, since the risk over states plus ambiguity is an upper bound of the EFE, minimising the upper bound will also minimise the EFE.

\subsubsection{Expected energy vs entropy formulation}

Finally, restarting from the risk over states plus ambiguity in Equation \ref{eq:risk_over_states_plus_ambiguity_in_setting_2}, one can demonstrate that:
\begin{align*}
	\mathcal{G}_{rt}(\overline{a}) &\leq \mathcal{C}_{RSA}(\overline{a}) = - \underbrace{H[F(\overline{s} | \overline{a})]}_{\text{entropy}} - \underbrace{\mathbb{E}_{F(\overline{o}, \overline{s} | \overline{a})} [\ln T(\overline{o}, \overline{s} | \overline{a})]}_{\text{expected energy}} = \mathcal{C}_{3E}(\overline{a}) \phantom{\sum}
\end{align*}
Thus, the expected energy vs entropy formulation can be recovered in this setup.
\vspace{0.5cm}
\begin{mdframed}[style=proof]
	Restarting from Equation \ref{eq:risk_over_states_plus_ambiguity_in_setting_2}:
	\begin{align*}
		\mathcal{G}_{rt}(\overline{a}) &\leq \mathcal{C}_{RSA}(\overline{a}) = \kl{F(\overline{s} | \overline{a})}{T(\overline{s} | \overline{a})} + \mathbb{E}_{F(\overline{s} | \overline{a})} \big[H[F(\overline{o} | \overline{s})]\big]. \phantom{\sum}
	\end{align*}
	and using the KL-divergence and entropy definitions, we obtain:
	\begin{align*}
		\mathcal{G}_{rt}(\overline{a}) \leq \mathcal{C}_{RSA}(\overline{a}) &= \mathbb{E}_{F(\overline{s} | \overline{a})} \big[ \ln F(\overline{s} | \overline{a}) - \ln T(\overline{s} | \overline{a}) \big] - \mathbb{E}_{F(\overline{o}, \overline{s} | \overline{a})} \big[\ln F(\overline{o} | \overline{s})\big]. \phantom{\sum}
	\end{align*}
	Then, using our assumption that $F(\overline{o} | \overline{s}) = T(\overline{o} | \overline{s})$, as well as the linearity of expectation and the log-property, we get:
	\begin{align*}
		\mathcal{G}_{rt}(\overline{a}) \leq \mathcal{C}_{RSA}(\overline{a}) &= \mathbb{E}_{F(\overline{s} | \overline{a})} \big[ \ln F(\overline{s} | \overline{a}) - \ln T(\overline{s} | \overline{a}) \big] - \mathbb{E}_{F(\overline{o}, \overline{s} | \overline{a})} \big[\ln T(\overline{o} | \overline{s})\big] \phantom{\sum}\\
		&= \mathbb{E}_{F(\overline{s} | \overline{a})} [\ln F(\overline{s} | \overline{a})] - \mathbb{E}_{F(\overline{o}, \overline{s} | \overline{a})} [\ln T(\overline{o}, \overline{s} | \overline{a})]. \phantom{\sum}
	\end{align*}
	Finally, recognizing the entropy definition, we obtain the desired result:
	\begin{align*}
		\mathcal{G}_{rt}(\overline{a}) \leq \mathcal{C}_{RSA}(\overline{a}) &= - \underbrace{H[F(\overline{s} | \overline{a})]}_{\text{entropy}} - \underbrace{\mathbb{E}_{F(\overline{o}, \overline{s} | \overline{a})} [\ln T(\overline{o}, \overline{s} | \overline{a})]}_{\text{expected energy}} = \mathcal{C}_{3E}(\overline{a}). \phantom{\sum}
	\end{align*}
\end{mdframed}

\section{Limitations}

In the previous section, we defined the expected free energy as the risk over observations plus ambiguity, i.e., $\mathcal{C}_{ROA}(\overline{a})$, and showed that it is equal to the information gain and pragmatic value formulation, i.e., $\mathcal{C}_{IGPV}(\overline{a})$. Then, following the proof in Appendix B of \cite{parr2022active}, the risk over states plus ambiguity $\mathcal{C}_{RSA}(\overline{a})$ was shown to be an upper bound of $\mathcal{C}_{ROA}(\overline{a})$. Finally, the entropy plus expected energy formulation $\mathcal{C}_{3E}(\overline{a})$ was derived from $\mathcal{C}_{RSA}(\overline{a})$. In summary:
\begin{align}
	\mathcal{G}_{rt}(\overline{a}) &\delequal \mathcal{C}_{ROA}(\overline{a}) = \mathcal{C}_{IGPV}(\overline{a}) \leq \mathcal{C}_{RSA}(\overline{a}) = \mathcal{C}_{3E}(\overline{a}). \label{eq:summary_efe_formulations}
\end{align}
Importantly, the proofs leading to equation \eqref{eq:summary_efe_formulations} rely on the assumption that the likelihood of the forecast and target distributions are equal, i.e., $F(\overline{o} | \overline{s}) = T(\overline{o} | \overline{s})$. In the following subsections, we study the limitations of the formalism presented in Section \ref{sec:planning}.

\subsection{Prior preferences over observations}

In this section, we study the assumptions made in Section \ref{ssec:solving_the_unification_problem} and their consequences. For simplicity, we only consider the case where the time horizon $h$ is equal to $t + 1$, and where $o_{t+1}$, $s_{t+1}$, as well as $a_t$ are discrete random variables. In this case, $\overline{o} = o_{t+1:h} = o_{t+1}$, $\overline{s} = s_{t+1:h} = s_{t+1}$, and similarly $\overline{a} = a_{t:h-1} = a_t$. Note, in Section \ref{ssec:solving_the_unification_problem}, the EFE is defined as the risk over observations plus ambiguity, i.e.,
\begin{align*}
	\mathcal{G}_{rt}(a_t) &\delequal \underbrace{\kl{F(o_{t+1} | a_t)}{T(o_{t+1} | a_t)}}_{\text{risk over observations}} + \underbrace{\mathbb{E}_{F(s_{t+1} | a_t)} \big[H[F(o_{t+1} | s_{t+1})]\big]}_{\text{ambiguity}} = \mathcal{C}_{ROA}(a_t).
\end{align*}

\subsubsection{The assumptions seemingly lead to an equation with no valid solution}

In this section, we show that the assumptions made in Section \ref{sec:planning}, seemingly lead to an equation with no valid solution. First, note that in the active inference literature, the prior preferences over observations are defined as follows: $T(o_{t+1} | a_t) = \text{Cat}(o_{t+1};\bm{C}_o)$. Additionally, recall that the proof in Section \ref{subsubsection:RSA_formulation_second_setting} relies on the assumption that $T(o_{t+1} | s_{t+1}) = F(o_{t+1} | s_{t+1})$, where the likelihood of the forecast distribution is also the likelihood of the generative model, i.e., $T(o_{t+1} | s_{t+1}) = F(o_{t+1} | s_{t+1}) = P(o_{t+1} | s_{t+1})$. In the active inference literature, the likelihood of the generative model is defined as: $P(o_{t+1} | s_{t+1}) = \text{Cat}(o_{t+1} | s_{t+1}; \bm{A})$. To sum up, we have $T(o_{t+1} | a_t) = \text{Cat}(o_{t+1};\bm{C}_o)$ and $T(o_{t+1} | s_{t+1}) = \text{Cat}(o_{t+1} | s_{t+1}; \bm{A})$. Using the sum and product rules of probability, we have:
\begin{align}
	T(o_{t+1} | a_t) = \sum_{s_{t+1}} T(o_{t+1} | s_{t+1}, a_t)T(s_{t+1} | a_t) \Leftrightarrow \bm{C}_o = \bm{A}\bm{C}_s, \label{eq:inconsistency_of_the_assumptions}
\end{align}
where without loss of generality, we let $T(s_{t+1} | a_t) = \text{Cat}(s_{t+1}; \bm{C}_s)$. Importantly, the above assumes that:
$$T(o_{t+1} | s_{t+1}, a_t) = T(o_{t+1} | s_{t+1}),$$
which holds because in the target distribution, the observations are independent of the policy given the states, i.e., using the d-separation criteria one can show that: $o_{t+1} \indep a_t \,\, | \,\, s_{t+1}$. Re-starting from \eqref{eq:inconsistency_of_the_assumptions}, one can solve for $\bm{C}_s$ and get:
\begin{align*}
	\bm{C}_o = \bm{A}\bm{C}_s \Leftrightarrow \bm{C}_s = \bm{A}^{-1}\bm{C}_o.
\end{align*}
However, the above equation may not have any valid solution. For example, if:
\begin{align}
	\bm{A} = \begin{bmatrix}
		0.6 & 0.4\\
		0.4 & 0.6
	\end{bmatrix}\,\,\text{ and }\,\,\bm{C}_o = \begin{bmatrix}
		0.8\\
		0.2
	\end{bmatrix}, \label{eq:a_matrix_appendix_b_1}
\end{align}
then one can show that:
\begin{align*}
	\bm{A}^{-1} = \begin{bmatrix}
		3 & -2\\
		-2 & 3
	\end{bmatrix}\,\,\text{ and }\,\,\bm{C}_s = \bm{A}^{-1}\bm{C}_o = \begin{bmatrix}
	2\\
	-1
	\end{bmatrix},
\end{align*}
which is not a valid solution as $\bm{C}_s$ are the parameters of a categorical distribution. Thus, the elements of $\bm{C}_s$ should add up to one, and be between zero and one. Importantly, finding a renormalization $\bar{\bm{C}}_s$ of $\bm{C}_s$ is impossible. Indeed, if $\bm{A}$ is invertible, then the inverse is unique. Thus, $\bm{C}_s$ is the only matrix that satisfies: $\bm{C}_s = \bm{A}^{-1}\bm{C}_o$.

\subsubsection{The class of valid prior preferences over observations}

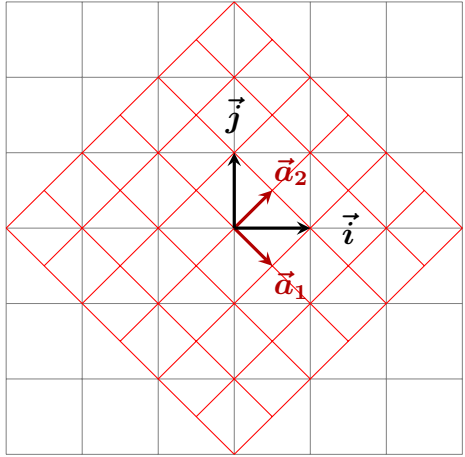
\begin{figure}[h]
	\begin{center}
		\begin{tikzpicture}
			\draw [step=1.0, gray, very thin] (-3,-3) grid (3,3);
			\draw [cm={0.5,0.5,0.5,-0.5,(0,0)}, red, thin, step=1.0cm] (-3,-3) grid (3,3);
			
			\draw [cm={0.5,0.5,0.5,-0.5,(0,0)}, -stealth, very thick, red!70!black] (0, 0) -- (0,1);
			\draw [cm={0.5,0.5,0.5,-0.5,(0,0)}, -stealth, very thick, red!70!black] (0, 0) -- (1,0);
			\node [red!70!black] at (0.75,0.75) {$\bm{\vec{a}_2}$};
			\node [red!70!black] at (0.75,-0.75) {$\bm{\vec{a}_1}$};
			
			\draw [-stealth, very thick] (0, 0) -- (0,1);
			\draw [-stealth, very thick] (0, 0) -- (1,0);
			\node at (0,1.5) {$\bm{\vec{j}}$};
			\node at (1.5,0) {$\bm{\vec{i}}$};
		\end{tikzpicture}
	\end{center}
	\vspace{-0.25cm}
	\caption{
		This figure illustrates how matrices can be seen as linear transformations of the Euclidean space. The example taken is a matrix that rotates the Euclidean space by 45 degree clockwise and scales each axis by a factor of 0.5.
	}
	\label{fig:matrix_as_linear_transformation}
\end{figure}

The problem described in the previous section occurs because we are defining two distributions over the random variable $o_{t+1}$, and these distributions are not compatible with each other. This indicates that in the setting of Section \ref{sec:planning}, one cannot defined the prior preferences over observations arbitrarily. Instead, given the likelihood mapping $T(o_{t+1} | s_{t+1}) = \text{Cat}(o_{t+1} | s_{t+1}; \bm{A})$, one can only define prior preferences over states, i.e., $T(s_{t+1} | a_t) = \text{Cat}(s_{t+1}; \bm{C}_s)$, Then, the prior preferences over observations can be computed as follows:
\begin{align}
	T(o_{t+1} | a_t) = \sum_{s_{t+1}} T(o_{t+1} | s_{t+1}, a_t)T(s_{t+1} | a_t) \Leftrightarrow \bm{C}_o = \bm{A}\bm{C}_s.
\end{align}
While this tells us how the prior over observations can be computed, it does not characterise the class of all valid prior preferences over observations. To characterise this class, we need to remember that $n \times n$ matrices can be understood as linear transformations of the $n$-dimensional Euclidean space. To understand this, let's take the following $2 \times 2$ matrix as an example:
\begin{align*}
	\bm{B} = \begin{bmatrix}
		\,\,\, \vert & \,\, \vert \,\,\, \\
		\,\,\, \vec{b}_1 & \,\, \vec{b}_2 \,\,\, \\
		\,\,\, \vert & \,\, \vert \,\,\, 
	\end{bmatrix},
\end{align*}
and let:
\begin{align*}
	\vec{i} = \begin{bmatrix}
		1 \\
		0
	\end{bmatrix},\text{ and: }\vec{j} = \begin{bmatrix}
		0 \\
		1
	\end{bmatrix},
\end{align*}
be the two standard basis vectors. One can see that: $\vec{b}_1 = \bm{B}\vec{i}$ and $\vec{b}_2 = \bm{B}\vec{j}$. In other words, the matrix $\bm{B}$ transforms the vector $\vec{i}$ into the vector $\vec{b}_1$, and the vector $\vec{j}$ into the vector $\vec{b}_2$. More generally, the matrix $\bm{B}$ transforms each vector $\vec{x}$ into a vector $\vec{y} = \bm{B}\vec{x}$. Geometrically, this can be understood as mapping each point of the Euclidean space $\vec{x}$, to another point $\vec{y}$ in the space. For example, the transformation corresponding to the following matrix:
\begin{align*}
	\bm{B} = \begin{bmatrix}
		0.5 & 0.5 \\
		0.5 & -0.5
	\end{bmatrix},
\end{align*}
is illustrated in Figure \ref{fig:matrix_as_linear_transformation}. Importantly, since the transformation is linear, the lines of the grid in Figure \ref{fig:matrix_as_linear_transformation} remain parallel and equally spaced. Thus, knowing where $\vec{i}$ and $\vec{j}$ land under the transformation $\bm{B}$ is enough to know where all the other points on the grid land, i.e., $\vec{i}$ and $\vec{j}$ defines one parallelogram and all the others (parallelograms) are obtained by copy and pasting this parallelogram along the transformed axes. Going back to the following equation: $\bm{C}_o = \bm{A}\bm{C}_s$, we can now better understand which prior preferences over observations $\bm{C}_o$ are compatible with the likelihood mapping $\bm{A}$. 

\begin{figure}[h]
	\begin{center}
		\begin{tikzpicture}
			\draw [step=1.0, gray, very thin] (-3,-3) grid (3,3);
			
			\draw [thick, blue!70!black] (1, 0) -- (0,1);
			\draw [dotted, blue!70!black] (3, -2) -- (-2,3);
			
			\draw [cm={0.6,0.4,0.4,0.6,(0,0)}, -stealth, very thick, red!70!black] (0, 0) -- (0,1);
			\draw [cm={0.6,0.4,0.4,0.6,(0,0)}, -stealth, very thick, red!70!black] (0, 0) -- (1,0);
			
			\draw [-stealth, very thick] (0, 0) -- (0,1);
			\draw [-stealth, very thick] (0, 0) -- (1,0);
			
			\node at (-1,2) [circle,fill,orange!60!black,inner sep=1.1pt] {};
			\node at (-0.8,2.2) [orange!60!black] {\ding{55}};
			
			\node at (0.5,0.5) [circle,fill,green!80!black,inner sep=1.1pt] {};
			\node at (0.7,0.7) [green!80!black] {\ding{51}};

			\draw [step=1.0, gray, very thin] (5,-3) grid (11,3);
			\draw [cm={0.6,0.4,0.4,0.6,(8,0)}, red, thin, step=1.0cm] (-3,-3) grid (3,3);
			
			\draw [cm={0.6,0.4,0.4,0.6,(8,0)}, thick, blue!70!black] (1, 0) -- (0,1);
			\draw [dotted, blue!70!black] (9, 0) -- (8,1);
			
			\draw [cm={0.6,0.4,0.4,0.6,(8,0)}, -stealth, very thick, red!70!black] (0, 0) -- (0,1);
			\draw [cm={0.6,0.4,0.4,0.6,(8,0)}, -stealth, very thick, red!70!black] (0, 0) -- (1,0);
			
			\draw [-stealth, very thick] (8, 0) -- (8,1);
			\draw [-stealth, very thick] (8, 0) -- (9,0);
			
			\node at (8.2,0.8) [circle,fill,orange!60!black,inner sep=1.1pt] {};
			\node at (8.4,1.0) [orange!60!black] {\ding{55}};
			
			\node at (8.5,0.5) [circle,fill,green!80!black,inner sep=1.1pt] {};
			\node at (8.7,0.7) [green!80!black] {\ding{51}};
			
		\end{tikzpicture}
	\end{center}
	\vspace{-0.25cm}
	\caption{
		This figure illustrates the linear transformation corresponding to the $\bm{A}$ matrix in Equation \ref{eq:a_matrix_appendix_b_1}. The gray grid represents a set of points in the Euclidean space on which the linear transformation corresponding to the $\bm{A}$ matrix will be applied. The red grid represents where the gray grid lands when applying this linear transformation. The long (blue) segment on the left-hand-side of the figure corresponds to the 1-dimensional simplex in which the prior preferences over states $\bm{C}_s$ lives. The short (blue) segment on the right-hand-side of the figure corresponds to where the 1-dimensional simplex lands when applying the linear transformation, i.e., the class of valid prior preferences over observations $\bm{C}_o$. The green point (\ding{51} on the right-hand-side) corresponds to a point that lives on the projected simplex (i.e., the short blue segment). This point will be projected back to the original 1-dimensional simplex (i.e., the long blue segment) by $\bm{A}^{-1}$. The brown point (\ding{55} on the right-hand-side) corresponds to a point that lives outside the short blue segment. This point will be projected outside of the long blue segment by $\bm{A}^{-1}$.
	}
	\label{fig:understanding_the_class_of_valid_prior_preferences}
\end{figure}
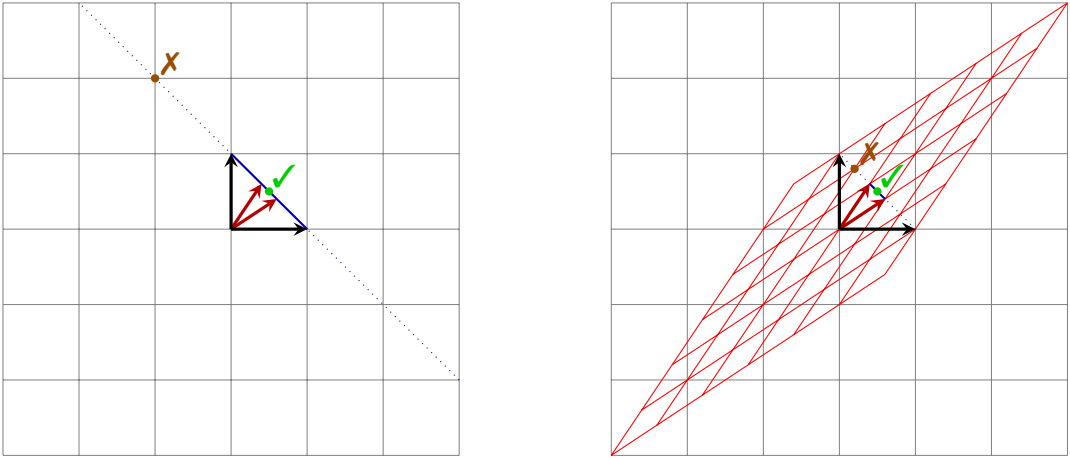

\noindent Note that $\bm{C}_o$ is a linear transformation of $\bm{C}_s$, where the transformation is defined by the elements of the matrix $\bm{A}$. Moreover, $\bm{C}_s$ are the parameters of a categorical distribution, which means that its elements are positive and sum up to one. Geometrically, this means that $\bm{C}_s$ is in the 1-dimensional simplex of the Euclidean space. Additionally, since $\bm{A}$ defines the probability of each observation given each state, all the elements of $\bm{A}$ are positive and the columns of $\bm{A}$ sum up to one. In other words, the columns of $\bm{A}$ correspond to points on the 1-dimensional simplex of the Euclidean space.
\\\\
\noindent Figure \ref{fig:understanding_the_class_of_valid_prior_preferences} illustrates the linear transformation corresponding to the $\bm{A}$ matrix of the previous section, c.f., Equation \ref{eq:a_matrix_appendix_b_1}. Recall that the columns of $\bm{A}$ (i.e., $\vec{a}_1$ and $\vec{a}_2$) correspond to points on the 1-dimensional simplex (represented in blue on the left of Figure \ref{fig:understanding_the_class_of_valid_prior_preferences}). Importantly, the standard basis vector $\vec{i}$ and $\vec{j}$ are mapped by $\bm{A}$ to $\vec{a}_1$ and $\vec{a}_2$, respectively. While this is happening, the gray grid (Figure \ref{fig:understanding_the_class_of_valid_prior_preferences} left) will be squeezed into the red grid (Figure \ref{fig:understanding_the_class_of_valid_prior_preferences} right), and the 1-dimensional simplex represented by a long blue segment will be squeezed into a shorter blue segment. This short blue segment is the class of valid prior preferences over observations $\bm{C}_o$.
\\\\
\noindent Indeed, the matrix $\bm{A}^{-1}$ performs the inverse linear transformation, i.e., $\bm{A}^{-1}$ transforms the red grid into the gray grid. Therefore, if a point is on the short blue segment in the right-hand-side of Figure \ref{fig:understanding_the_class_of_valid_prior_preferences}, it will be mapped back to the original 1-dimensional simplex (the long blue segment). However, if a point starts on the blue dotted line (outside the short blue segment), it will be mapped outside of the original 1-dimensional simplex. 
\\\\
\noindent To conclude, the class of valid prior preferences over observations is the class of all vectors $\bm{C}_o$ that can be obtained as a linear combination of the columns of $\bm{A}$, where the weights of the linear combination are positive numbers between zero and one that sum up to one, i.e., all the vectors that satisfies the equation $\bm{C}_o = \bm{A}\bm{C}_s$. Geometrically, the class of valid prior preferences over observations corresponds to the short blue segment obtained by applying $\bm{A}$ to the 1-dimensional simplex.

\subsection{Justification of the expected free energy} \label{ssec:justification}

In this section, we discuss the expected free energy justification presented in Appendix B of \cite{parr2022active}. The appendix derives $\mathcal{C}_{RSA}(\overline{a})$ from the assumption that an agent aims to reach its prior preferences, i.e., $F(s_\tau) = T(s_\tau | a_{\tau - 1})$ for some time step $\tau$ in the future. Importantly, this derivation provides a justification for $\mathcal{C}_{RSA}(\overline{a})$, i.e., an upper bound of $\mathcal{G}_{rt}(\overline{a})$ as defined in Section \ref{sec:planning}, but not for $\mathcal{G}_{rt}(\overline{a})$ itself. Thus, the expected free energy of Section \ref{sec:planning}, i.e., $\mathcal{C}_{ROA}(\overline{a})$, remains justified only by its intuitive definition. To resolve this issue, we could try to define the expected free energy as the risk over states plus ambiguity:
\begin{align*}
	\mathcal{C}_{ROA}(\overline{a}) = \mathcal{C}_{IGPV}(\overline{a}) \leq \mathcal{C}_{RSA}(\overline{a}) = \mathcal{C}_{3E}(\overline{a}) \delequal \mathcal{G}_{rt}(\overline{a}).
\end{align*}
In this case, we have a justification for the expected free energy, i.e., $\mathcal{C}_{RSA}(\overline{a})$, but $\mathcal{C}_{ROA}(\overline{a})$ and $\mathcal{C}_{IGPV}(\overline{a})$ are now lower bounds of $\mathcal{G}_{rt}(\overline{a})$. As minimizing a lower bound of the expected free energy does not imply minimizing the expected free energy, $\mathcal{C}_{ROA}(\overline{a})$ and $\mathcal{C}_{IGPV}(\overline{a})$ cannot be recovered in this setting. The results of the derivations presented in this paper are summarized in Table \ref{tab:derivation_results_for_efe_formulations}. Out of two potential definitions for the expected free energy, one is justified but can only recover two EFE formulations, and the other is not currently justified but can recover the four formulations.

\begin{table}[h]
	\centering
	\begin{tabular}{c c c c c c c c c c c}
		& \mcrot{2}{l}{60}{$\mathcal{C}_{IGPG}$} & \mcrot{2}{l}{60}{$\mathcal{C}_{RSA}$} & \mcrot{2}{l}{60}{$\mathcal{C}_{ROA}$} & \mcrot{2}{l}{60}{$\mathcal{C}_{3E}$} & \mcrot{2}{l}{60}{justified} \\
		\hline
		risk over states vs ambiguity as definition &  \multicolumn{2}{l}{\ding{55}} & \multicolumn{2}{l}{\ding{51}} & \multicolumn{2}{l}{\ding{55}} & \multicolumn{2}{l}{\ding{51}} & \multicolumn{2}{l}{\ding{51}} \\
		\hline
		risk over observations vs ambiguity as definition &  \multicolumn{2}{l}{\ding{51}} & \multicolumn{2}{l}{\ding{51}} & \multicolumn{2}{l}{\ding{51}} & \multicolumn{2}{l}{\ding{51}} & \multicolumn{2}{l}{\ding{55}} \\
	\end{tabular}
	\caption{This table summarises the results of the derivations of the expected free energy formulations, and whether a justification of the EFE has been provided in the literature. The first row corresponds to the definition of \citet{parr2022active} where the expected free energy is defined as the risk over states vs ambiguity formulation, while the second row corresponds to another interpretation of \citet{parr2022active} where the expected free energy is defined as the risk over observations vs ambiguity formulation. Cells containing \ding{55} mean that the formulation cannot be recovered (or no justification is known), while cells containing \ding{51} correspond to the case where the formulation can be recovered (or a justification is known).}
	\label{tab:derivation_results_for_efe_formulations}
\end{table}

\section{Conclusion}

This paper aimed to formalize the expected free energy definition, as well as the problem of deriving its four formulations, i.e., the unification problem. When the expected free energy is defined as the risk over observations plus ambiguity, all formulations can be recovered, and can therefore be used in practice. However, an important contribution of this paper was to show that some prior preferences over observations are incompatible with the likelihood mapping. Thus, we are left with a dilemma, either the modeller has to carefully pick the prior preferences of the agent to avoid any conflict, or we have to let go of the theoretical connection between the four formulations.

Another issue is the absence of a justification for the risk over observations plus ambiguity formulation. While there exists a justification for the risk over states plus ambiguity formulation, justifying a lower bound is not enough to justify the expected free energy. Therefore, future research should focus on finding a derivation of the risk over observations plus ambiguity from first principles. Importantly, while the risk over states plus ambiguity is justified, this definition of the expected free energy does not allow us to recover the four formulations. Thus, it does not constitute a valid solution to the unification problem.

Note, we only studied two possible definitions of the expected free energy. An alternative set of proofs and/or another factorization of the forecast and target definitions may allow us to recover all four decompositions, while also removing the conflict between prior preferences and likelihood. However, testing all possible factorizations and proofs is outside the scope of this paper.

Finally, this paper provides a solid foundation for future research, especially in the context of deep active inference. Indeed, this paper clarifies the expected free energy definition, but unfortunately, it does explain how to compute it using deep neural networks. Thus, additional research is required to implement and empirically evaluate the proposed expected free energy definition.

\vskip 0.2in
\bibliography{efe_formulation.bib}

\section*{Appendix A: comments, questions, answers, and proofs}

In this paper, we used the following blocks:
\vspace{0.5cm}
\begin{mdframed}[style=comment]
	A comment.
\end{mdframed}
\begin{mdframed}[style=question]
	A question.
\end{mdframed}
\begin{mdframed}[style=answer]
	An answer.
\end{mdframed}
\begin{mdframed}[style=proof]
	A proof.
\end{mdframed}

\section*{Appendix B: important properties}

In this paper, we made extensive use of four basic properties. The first is the sum-rule of probability. Given two sets of random variables $\bm{X}$ and $\bm{Y}$, the sum-rule of probability states that:
\begin{align}
	P(\bm{Y}) = \int_{\bm{X}} P(\bm{X},\bm{Y}) \,\, d\bm{X}.
\end{align}
The sum-rule can then be used to sum out random variables from a joint distribution. The second property is called the product-rule of probability, and can be used to split a joint distribution into conditional distributions. Given two sets of random variables $\bm{X}$ and $\bm{Y}$, the product-rule of probability states that:
\begin{align}
	P(\bm{X},\bm{Y}) = P(\bm{X}|\bm{Y}) P(\bm{Y}).
\end{align}
Next, if $\bm{a}$ and $\bm{b}$ are two real numbers, then a relevant property of logarithm is the following: 
\begin{align}
	\ln(\bm{a} \times \bm{b}) = \ln(\bm{a}) + \ln(\bm{b}).
\end{align}
Put simply, this allows us to turn the logarithm of a product into a sum of logarithms, and we will refer to the above equation as the ``log-property". Finally, the last property is called the linearity of expectation. Given a random variable $\bm{X}$, and two real numbers $\bm{a}$ and $\bm{b}$, the linearity of expectation states that:
\begin{align}
	\mathbb{E}[\bm{a}\bm{X} + \bm{b}] = \bm{a}\mathbb{E}[\bm{X}] + \bm{b},
\end{align}
where the expectation is w.r.t. the marginal distribution over $\bm{X}$, i.e., $P(\bm{X})$.

\section*{Appendix C: d-separation criterion}

Given a Bayesian network, the d-separation criterion provides a bridge between the graph topology and the independence assumptions holding within the Bayesian network. Let $\mathcal{G}_{rt} = (\mathcal{V}, \mathcal{E})$ be a directed graph corresponding to a Bayesian network, where $\mathcal{V}$ and $\mathcal{E}$ are the graph's vertices and edges, respectively. A trail is a sequence of vertices $(V_1, V_2, ..., V_k)$ such that there is an edge $V_i \rightarrow V_{i+1}$ or $V_{i+1} \rightarrow V_{i}$ for all $i \in \{1, ..., k - 1\}$.	Intuitively, trails connect two variables $V_1$ and $V_k$; conceptually if a trail is blocked, then $V_1$ does not provide any new information about $V_k$ through this trail. The notion of blocked trail is based on colliders:

\begin{definition}[Collider]
	Within a trail $(V_1, V_2, ..., V_k)$, a collider is a vertex $V_j$ s.t. $V_{j-1} \rightarrow V_j \leftarrow V_{j+1}$.
\end{definition}

\begin{definition}[Blocked trail]
	Given a set of vertices $\bm{S} \subseteq \mathcal{V}$, and two vertices $V_1, V_2 \in \mathcal{V}$, we say that a trail between $V_1$ and $V_2$ is blocked by $\bm{S}$ if at least one node of the trail is a collider not in $\bm{S}$ and with no descendants in $\bm{S}$, or at least one vertex of the trail that is not a collider is in $\bm{S}$.
\end{definition}

\noindent Importantly, two vertices in the graph can be connected through multiple trails; if all trails between these two vertices are blocked, then we say that these vertices are d-separated. This can be generalized to sets of vertices as shown below:

\begin{definition}[D-separated]
	Given a set of vertices $\bm{S} \subseteq \mathcal{V}$, and two vertices $V_1, V_2 \in \mathcal{V}$, we say that $V_1$ and $V_2$ are d-separated by $\bm{S}$ if all trails between $V_1$ and $V_2$ are blocked.
\end{definition}

\vspace{0.1cm}
\begin{mdframed}[style=comment]
	Given three sets of vertices $\bm{V}_1, \bm{V}_2, \bm{S} \subseteq \mathcal{V}$. We say that $\bm{V}_1$ and $\bm{V}_2$ are d-separated by $\bm{S}$, if each node in $\bm{V}_1$ is d-separated from all nodes in $\bm{V}_2$ given the nodes in $\bm{S}$.
\end{mdframed}
\vspace{0.25cm}

\noindent Finally, the d-separation theorem states that: if two sets of vertices are d-separated in the graph, then the associated random variables are conditionally independent, i.e.,

\begin{theorem}[d-separation and independence]
	Given three sets $\bm{V}_1, \bm{V}_2, \bm{S} \subseteq \mathcal{V}$. If $\bm{V}_1$ and $\bm{V}_2$ are d-separated by $\bm{S}$, then $\bm{V}_1$ and $\bm{V}_2$ are conditionally independent given $\bm{S}$, i.e., $\bm{V}_1 \indep \bm{V}_2 \mid \bm{S}$.
\end{theorem}

\noindent In practice, the d-separation theorem is used on the generative model's graph. The goal is to know whether a conditional assumption holds, i.e., whether $\bm{V}_1 \indep \bm{V}_2 \mid \bm{S}$ holds. If it does, then: $P(\bm{V}_1 | \bm{S}, \bm{V}_2) = P(\bm{V}_1 | \bm{S})$.

\end{document}